\documentclass{article}

\usepackage{arxiv}

\usepackage[utf8]{inputenc}
\usepackage[T1]{fontenc}
\usepackage[dvipsnames]{xcolor}
\usepackage{hyperref}
\usepackage{url}
\usepackage{booktabs}
\usepackage{multirow}
\usepackage{tabularx}
\usepackage{array}
\usepackage{amsmath,amssymb,amsfonts}
\usepackage{bm}
\usepackage{microtype}
\usepackage{graphicx}
\usepackage[numbers,square,sort&compress]{natbib}
\usepackage{cleveref}
\usepackage{caption}
\newcommand{\Rsq}{R\textsuperscript{2}}
\newcolumntype{Y}{>{\raggedright\arraybackslash}X}

\title{Calibrating Prediction Timeliness Through Multi-Objective Hyperparameter Optimization for Remaining Useful Life Prediction}

\newbox{\orcid}\sbox{\orcid}{\includegraphics[scale=0.06]{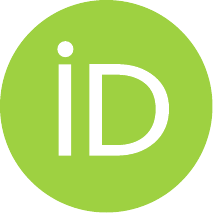}}

\author{%
  \href{https://orcid.org/0000-0003-0646-8950}{\usebox{\orcid}\hspace{1mm}Tu\u{g}rul Cabir Hakyemez}\thanks{Corresponding author.} \\
  Department of Management Information Systems\\
  Istanbul Bilgi University\\
  \.{I}stanbul, Turkey \\
  \texttt{tugrul.hakyemez@bilgi.edu.tr} \\
  \And
  Ener Uras G\"{o}khan \\
  Be\c{s}tepe Koleji\\
  Ankara, Turkey \\
  \texttt{eneruras@gmail.com} \\
}

\renewcommand{\shorttitle}{Multi-Objective HPO for RUL Prediction}

\hypersetup{
  colorlinks=true, linkcolor=blue!60!black, citecolor=blue!60!black, urlcolor=blue!60!black,
  pdftitle={Calibrating Prediction Timeliness Through Multi-Objective Hyperparameter Optimization for Remaining Useful Life Prediction},
  pdfauthor={Tugrul Cabir Hakyemez, Ener Uras Gokhan},
  pdfkeywords={Remaining Useful Life, hyperparameter optimization, multi-objective optimization, NSGA-II, Entropy-CRITIC, predictive maintenance, C-MAPSS, BackBlaze},
}

\begin{document}
\maketitle

\begin{abstract}
In predictive maintenance, early and late RUL prediction errors carry asymmetric consequences, yet hyperparameter optimization typically targets a single accuracy metric that treats both directions equally. This study treats the optimization objective itself as a design variable. Five architectures---MLP, LSTM, XGBoost, TCN, and Transformer---are evaluated under three regimes: single-objective maximization of \Rsq{}, single-objective minimization of the NASA scoring function, and a multi-objective formulation that jointly optimizes both criteria. The multi-objective search employs NSGA-II with Entropy--CRITIC weighting for Pareto selection. Seventy-five model--dataset--strategy combinations are assessed on the NASA C-MAPSS turbofan and BackBlaze hard-disk drive benchmarks. On C-MAPSS, all strategies achieve comparable accuracy (\Rsq{}~$\approx$~0.89), yet multi-objective optimization reduces directional imbalance by approximately 33\%, improving calibration of early versus late predictions. Model rankings prove configuration-dependent, with simpler architectures frequently outperforming deeper temporal models. On BackBlaze, the objectives shift from complementary to conflicting, producing divergent Entropy--CRITIC weights and a substantial generalization gap (best \Rsq{}~$\approx$~0.34). These results demonstrate that the optimization objective materially shapes prognostic behavior and that multi-objective search provides a practical mechanism for calibrating prediction timeliness in RUL modeling.
\end{abstract}

\keywords{Remaining Useful Life \and hyperparameter optimization \and multi-objective optimization \and NSGA-II \and Entropy--CRITIC \and predictive maintenance \and C-MAPSS \and BackBlaze}

\section{Introduction}

In predictive maintenance, early and late Remaining Useful Life (RUL) prediction errors carry fundamentally asymmetric consequences: late predictions risk catastrophic damage, safety incidents, and unplanned downtime, whereas early predictions primarily incur premature maintenance costs~\cite{r1,r2,r3,r4}. Yet standard regression metrics such as RMSE and \Rsq{} treat both directions equally, creating a misalignment between optimization objectives and operational costs~\cite{r4,r5}. This mismatch motivates asymmetric alternatives---notably the NASA Prognostics Score, which penalizes late predictions more heavily~\cite{r4,r6,r7,r8,r9}---and raises a deeper question: how does the choice of optimization objective itself shape the resulting prognostic behavior?

Hyperparameter optimization (HPO) is a critical determinant of RUL model performance~\cite{r10,r11,r12}, yet it is predominantly conducted under a single accuracy-based criterion~\cite{r13,r14}. Multi-objective optimization (MOO) offers an alternative by simultaneously addressing conflicting goals, but prior MOO-based RUL studies typically balance prediction error against computational cost without a principled final selection procedure~\cite{r13,r15}. Whether single- and multi-objective regimes produce meaningfully different prognostic profiles---beyond aggregate accuracy---remains largely unexamined~\cite{r14,r17}. Furthermore, the predominant reliance on the simulated C-MAPSS benchmark~\cite{r9} limits understanding of how optimization choices interact with real-world degradation dynamics, where noise and heterogeneous failure modes may alter the relationship between competing objectives~\cite{r6,r12,r18}.

This study treats the optimization objective as a first-class design variable. Five machine learning architectures---MLP, XGBoost, LSTM, TCN~\cite{r19}, and Transformer~\cite{r20}---are evaluated under three HPO regimes: single-objective \Rsq{} maximization, single-objective NASA Score minimization, and a multi-objective formulation that jointly optimizes both criteria using NSGA-II~\cite{r21} with Entropy--CRITIC post-hoc selection~\cite{r22} within the Optuna framework~\cite{r23}. A total of 75 model--dataset--strategy combinations are evaluated on the NASA C-MAPSS turbofan benchmark~\cite{r24} and the BackBlaze hard disk drive dataset~\cite{r25,r26}, using a multi-metric framework encompassing point accuracy (\Rsq{}, RMSE), safety-critical scoring (NASA Score)~\cite{r27}, precision ($\alpha$-accuracy at 20\%)~\cite{r28}, and timeliness indicators (Directional Imbalance).

The contributions of this work are threefold:
\begin{itemize}
  \item Five architectures are compared under identical single- and multi-objective optimization regimes, isolating the effect of the optimization objective on prognostic behavior.
  \item Entropy--CRITIC weighting provides principled Pareto front selection without subjective preference articulation.
  \item Cross-domain validation on simulated and real-world benchmarks addresses the data-fidelity gap in the RUL literature.
\end{itemize}

\section{Literature Review}

\subsection{RUL Prediction}

Remaining Useful Life (RUL) is defined as the duration a component can still perform its intended function before failing, serving as a cornerstone of Prognostics and Health Management (PHM)~\cite{r3,r29}. RUL prediction enables Condition-Based Maintenance (CBM) by utilizing historical and real-time monitoring data to prevent catastrophic failures and optimize operational safety~\cite{r30}.

Methodologies for estimating RUL are categorized into model-based, data-driven, and hybrid approaches~\cite{r3}. Within the data-driven domain, techniques have progressed from traditional machine learning to deep learning architectures---CNNs, LSTMs, attention mechanisms---capable of capturing complex spatial and temporal dependencies in high-dimensional sensor data~\cite{r4,r11,r32}.

The evaluation of RUL models relies predominantly on simulated benchmarks such as C-MAPSS, which provide complete run-to-failure trajectories but frequently fail to reproduce the full complexity of real operating conditions~\cite{r9}. True operational datasets remain scarce---accounting for only approximately 24\% of the data used in recent literature~\cite{r33}---raising concerns about the generalizability of models trained exclusively on simulated data.

\subsection{HPO in RUL Prediction}

Hyperparameter optimization (HPO) is a fundamental component of the RUL prediction pipeline, as parameters such as layer depth, batch size, and learning rate significantly influence a model's ability to learn degradation patterns and generalize to unseen data~\cite{r10,r13}. The relationship between hyperparameters and generalization error is typically non-convex and non-differentiable, necessitating advanced search strategies such as Bayesian optimization or evolutionary algorithms~\cite{r14}.

Single-objective optimization (SOO) approaches focus on a lone accuracy metric, treating RUL estimation as a standard curve-fitting problem~\cite{r4,r6,r10}, while neglecting operational factors such as asymmetric error costs~\cite{r15}. Multi-objective optimization (MOO) frameworks address multiple conflicting objectives simultaneously using algorithms like NSGA-II, seeking Pareto-optimal solutions where no objective can be improved without degrading another~\cite{r13,r36,r37}.

\subsection{This Study}

This study systematically compares three optimization strategies---single-objective \Rsq{} maximization, single-objective NASA Score minimization~\cite{r27}, and multi-objective optimization via NSGA-II jointly targeting both---across five architectures (MLP, LSTM, XGBoost, TCN, Transformer) and two cross-domain datasets: C-MAPSS and BackBlaze. The multi-objective variant is augmented with Entropy--CRITIC post-hoc weighting for principled Pareto front selection, eliminating the ad-hoc decision step that limits existing MOO-based studies~\cite{r13,r37}. All experiments follow a strict leakage-free temporal validation protocol.

As \Cref{tab:related} summarizes, existing studies typically optimize a single architecture on one simulated benchmark using symmetric error metrics. This study departs from prior work by: (i)~benchmarking five heterogeneous architectures under identical optimization and evaluation conditions, (ii)~introducing Entropy--CRITIC weighting as a reproducible selection layer atop the Pareto front, (iii)~adopting an evaluation suite combining standard metrics with asymmetric scoring and directional indicators~\cite{r28}, and (iv)~validating across simulated and real-world datasets, directly addressing the data-fidelity gap in the literature~\cite{r33}.

\begin{table}[t]
  \centering
  \caption{Key related research.}
  \label{tab:related}
  \scriptsize
  \setlength{\tabcolsep}{4pt}
  \renewcommand{\arraystretch}{1.25}
  \begin{tabularx}{\textwidth}{@{}p{1.7cm}YYYYY@{}}
    \toprule
    \textbf{Study} & \textbf{Dataset} & \textbf{Prediction algorithm(s)} & \textbf{Optimization algorithm} & \textbf{Optimization strategy} & \textbf{Evaluation metrics} \\
    \midrule
    Haris et al.~\cite{r10} & Supercapacitor dataset (simulated) & Deep Belief Network & Bayesian Optimization and Hyperband (BOHB) & Minimize RMSE & RMSE, MAE, \Rsq{} \\
    Li \& He~\cite{r5} & C-MAPSS & DCNN & Bayesian Optimization & Minimize the Deep Learning Error (DLE) & RMSE, Score \\
    Kong et al.~\cite{r11} & PCoE & CNN + LSTM & Bayesian Optimization & Minimize RMSE & MAE, MAPE, RMSE \\
    Elsherif et al.~\cite{r48} & C-MAPSS & CAELSTM & Tree-structured Parzen Estimator (TPE) & Minimize RMSE & RMSE, MAE, NASA Score \\
    Yang et al.~\cite{r49} & NASA battery dataset, CALCE & GBLS Booster & Tree-structured Parzen Estimator (TPE) & Minimize MAPE & MAPE, RMSE, \Rsq{}, Relative Error (RE) \\
    Kara~\cite{r50} & NASA battery dataset & CNN-LSTM & PSO & Minimize RMSE & MAE, MAPE, RMSE, AE \\
    Liu et al.~\cite{r51} & NASA battery dataset, CALCE & Mamba & Sinusoidal and Cauchy Sparrow Search Algorithm (SCSSA) & Minimize RE & RE, MAE, RMSE \\
    \midrule
    \textbf{This study} & \textbf{C-MAPSS, BackBlaze hard drives} & \textbf{MLP, LSTM, XGBoost, TCN, Transformer} & \textbf{NSGA-II, TPE} & \textbf{Minimize NASA, maximize \Rsq{}, both (multi-objective) + Entropy--CRITIC} & \textbf{\Rsq{}, RMSE, NASA Score, $\alpha$-20\%, DI} \\
    \bottomrule
  \end{tabularx}
\end{table}

\section{Methodology}

This section presents the experimental methodology employed to compare single-objective and multi-objective hyperparameter optimization strategies for RUL prediction. We describe the benchmark datasets, the candidate algorithms, the optimization frameworks, and the evaluation metrics. Our experimental design draws on established prognostics benchmarks and evaluation protocols from the PHM literature~\cite{r24,r27} while extending them through a systematic multi-objective optimization perspective.

\subsection{Datasets}

\subsubsection{C-MAPSS}

The Commercial Modular Aero-Propulsion System Simulation (C-MAPSS) dataset~\cite{r24} contains run-to-failure degradation trajectories of turbofan engines simulated under varying operating conditions and fault modes. The dataset comprises four subsets of increasing complexity: FD001 (single operating condition, single fault mode), FD002 (six operating conditions, single fault mode), FD003 (single operating condition, two fault modes), and FD004 (six operating conditions, two fault modes). Each subset provides multivariate time-series readings from 14 sensors recorded at each operational cycle until engine failure, using the standard train and test text files provided by NASA. Following common practice~\cite{r38,r39}, we apply a piecewise linear degradation model that caps the maximum RUL at 125 cycles. The training partitions are further split into training and validation sets for hyperparameter selection: FD001 and FD003 each contain 85 training and 15 validation engines with 100 test engines, while FD002 contains 221 training and 39 validation engines with 259 test engines, and FD004 contains 212 training and 37 validation engines with 248 test engines.

\subsubsection{BackBlaze}

The BackBlaze hard disk drive dataset~\cite{r25} provides daily SMART (Self-Monitoring, Analysis and Reporting Technology) attribute readings from operational data center drives. Unlike C-MAPSS, BackBlaze presents noisier sensor signals, less predictable degradation patterns, and more heterogeneous failure modes~\cite{r26}. We use data from three consecutive quarters (Q1--Q3 2024), merged with cross-quarter deduplication. Only drives with a recorded failure event are retained; censored (non-failed) drives are excluded. Drives with fewer than 30 days of operational history or fewer than 30 observations are filtered out to ensure sufficient degradation trajectory length. From the available SMART attributes, 14 features with established failure-predictive relevance are selected---including Reallocated Sectors Count, Reported Uncorrectable Errors, Current Pending Sector Count, Power-On Hours, and Temperature, among others. Each daily record is treated as one operational cycle, and the target RUL is computed as the number of days remaining until the recorded failure date. Analogous to the 125-cycle cap applied on C-MAPSS, RUL values are clipped at a maximum of 60 days, reflecting the assumption that drives far from failure occupy a functionally healthy regime. The resulting dataset comprises 2,444 failed drives (140,279 observations across 14 features), split at the drive level into training (1,956 drives; 111,971 observations) and test (488 drives; 28,308 observations) partitions with a fixed random seed to ensure reproducibility.

\Cref{tab:datasets} summarizes the key characteristics of all five datasets. The two benchmarks differ markedly in trajectory length (206--247 cycles on C-MAPSS vs.\ 57.6 days on BackBlaze) and in the proportion of observations at the RUL cap (39--49\% vs.\ 10.9\%), indicating that BackBlaze drives spend proportionally more of their observed life in active degradation yet under substantially noisier conditions.

\begin{table}[t]
  \centering
  \caption{Dataset summary.}
  \label{tab:datasets}
  \small
  \begin{tabular}{@{}lccccc@{}}
    \toprule
     & \textbf{FD001} & \textbf{FD002} & \textbf{FD003} & \textbf{FD004} & \textbf{BackBlaze} \\
    \midrule
    Train / Val / Test units   & 85 / 15 / 100 & 221 / 39 / 259 & 85 / 15 / 100 & 212 / 37 / 248 & 1,956 / --- / 488 \\
    Train observations         & 20,631 & 53,759 & 24,720 & 61,249 & 111,971 \\
    Test observations          & 13,096 & 33,991 & 16,596 & 41,214 & 28,308 \\
    Mean cycles/unit (train)   & 206.3 & 206.8 & 247.2 & 246.0 & 57.6 \\
    Mean train RUL             & 86.8 & 86.9 & 93.1 & 93.0 & 30.0 \\
    \% at RUL cap (train)      & 39.4\% & 39.5\% & 49.4\% & 49.2\% & 10.9\% \\
    \bottomrule
  \end{tabular}
\end{table}

\subsection{Candidate Algorithms}

We select five representative algorithms that span the spectrum of model complexity and architectural philosophy for sequential prognostics tasks. This diversity ensures that our conclusions regarding optimization strategies are not artifacts of a particular model family, following the recommendation of Lei et al.~\cite{r34} for comprehensive algorithmic comparison in prognostics research.

\subsubsection{Multilayer Perceptron (MLP)}

The MLP serves as a baseline feedforward architecture that processes flattened input windows without any explicit temporal inductive bias. Despite its simplicity, prior work has shown that well-tuned MLPs can be competitive with more complex architectures on certain prognostics benchmarks~\cite{r40}. Including the MLP allows us to assess whether the benefits of multi-objective optimization are contingent upon model complexity or whether simpler architectures can also leverage multi-objective search effectively.

\subsubsection{Long Short-Term Memory (LSTM)}

The LSTM network~\cite{r41} is a recurrent architecture specifically designed to capture long-range temporal dependencies through gated memory cells. It remains one of the most established deep learning approaches for RUL prediction~\cite{r42,r43}, owing to its ability to selectively retain and forget information across degradation trajectories. Zheng et al.~\cite{r42} were among the first to demonstrate the effectiveness of LSTM networks for direct RUL estimation on the C-MAPSS benchmark. We employ a stacked LSTM configuration with tunable depth, hidden dimension, and dropout regularization.

\subsubsection{XGBoost}

XGBoost~\cite{r44} is a gradient-boosted decision tree ensemble that operates on tabular feature representations. It provides a strong non-neural baseline that is robust to hyperparameter choices and typically requires less computational overhead than deep learning alternatives. Feature engineering for XGBoost includes statistical aggregates computed over sliding windows of the raw sensor data, enabling the model to capture temporal patterns indirectly. The inclusion of XGBoost follows the growing recognition that ensemble tree methods remain highly competitive across diverse machine learning tasks, including predictive maintenance applications~\cite{r45}.

\subsubsection{Temporal Convolutional Network (TCN)}

The TCN~\cite{r19} employs dilated causal convolutions to model temporal dependencies with a receptive field that grows exponentially with network depth, while maintaining parallelizability and stable gradient flow. Bai et al.~\cite{r19} demonstrated that generic convolutional architectures can match or outperform canonical recurrent networks across diverse sequence modeling tasks. Key hyperparameters include the number of residual blocks, kernel size, dilation factors, and channel dimensions.

\subsubsection{Transformer}

The Transformer architecture~\cite{r20} leverages multi-head self-attention to model pairwise interactions across all time steps in the input sequence. While originally developed for natural language processing, Transformers have shown considerable promise in time-series forecasting and prognostics tasks~\cite{r46,r47}. We adapt the encoder-only Transformer for RUL regression, with tunable attention heads, embedding dimension, feedforward dimension, and number of encoder layers. The self-attention mechanism enables the model to dynamically weight the relevance of different time steps in the degradation trajectory.

\subsection{Hyperparameter Optimization Strategies}

We compare three hyperparameter optimization strategies that differ in their objective formulation, all implemented within the Optuna framework~\cite{r23} using the Tree-structured Parzen Estimator (TPE) sampler.

\subsubsection{Single-Objective Optimization}

Two single-objective strategies are evaluated: maximization of \Rsq{} on the validation set, representing the standard approach in the prognostics literature~\cite{r35,r39}, and minimization of the NASA Prognostics Scoring Function~\cite{r27}, which applies exponentially asymmetric penalties that weigh late predictions more heavily than early ones. The sensitivity of the NASA Score to outlier predictions differentiates models tuned under this objective from those optimized for \Rsq{}.

\subsubsection{Multi-Objective Optimization}

\Rsq{} and NASA Score are optimized simultaneously via NSGA-II~\cite{r21}, both cast as minimization targets ($1 - R^2$ and NASA Score). From the resulting Pareto front $\mathcal{P}$ of $K$ solutions evaluated on $M = 2$ objectives, Entropy--CRITIC weighting~\cite{r22} selects a single configuration without subjective preference articulation. Objective values are min--max normalized into a decision matrix $\mathbf{R} = [r_{km}]$. Entropy weights capture discriminative information:
\begin{equation}
  w_m^{E} = \frac{1 - E_m}{\sum_{j}(1 - E_j)}, \qquad
  E_m = -\frac{1}{\ln K}\sum_{k=1}^{K} p_{km} \ln p_{km}.
\end{equation}
CRITIC weights incorporate inter-criteria contrast and correlation:
\begin{equation}
  w_m^{C} = \frac{C_m}{\sum_{j} C_j}, \qquad
  C_m = \sigma_m \sum_{j}\left(1 - \rho_{mj}\right),
\end{equation}
where $\sigma_m$ is the standard deviation of normalized objective $m$ and $\rho_{mj}$ the Pearson correlation between objectives. The final combined weights are
\begin{equation}
  w_m = \frac{w_m^{E} \cdot w_m^{C}}{\sum_{j} w_j^{E} \cdot w_j^{C}},
\end{equation}
used to select the Pareto-optimal solution with the highest composite score:
\begin{equation}
  k^{*} = \arg\max_{k \in \mathcal{P}} \sum_{m=1}^{M} w_m \cdot r_{km}.
\end{equation}
This data-driven selection is particularly appropriate when no domain expert is available to specify preference weights, as it derives weights from the information content and correlation structure of the Pareto front itself.

\subsection{Evaluation Metrics}

We adopt five complementary metrics that capture different facets of RUL prediction quality: point accuracy, safety-critical scoring, prediction precision, and directional bias. This multi-dimensional evaluation follows the recommendations of Saxena et al.~\cite{r28}, which emphasize that no single metric can fully characterize prognostic performance.

\subsubsection{Regression Accuracy}

Two standard regression metrics assess overall prediction accuracy. The coefficient of determination (\Rsq{}) quantifies explained variance:
\begin{equation}
  R^2 = 1 - \frac{\sum_{i=1}^{N}\left(y_i - \hat{y}_i\right)^2}{\sum_{i=1}^{N}\left(y_i - \bar{y}\right)^2},
\end{equation}
where $y_i$, $\hat{y}_i$, and $\bar{y}$ denote the true RUL, predicted RUL, and mean true RUL, respectively. The Root Mean Squared Error (RMSE) measures average prediction deviation with emphasis on large errors:
\begin{equation}
  \mathrm{RMSE} = \sqrt{\frac{1}{N}\sum_{i=1}^{N}\left(y_i - \hat{y}_i\right)^2}.
\end{equation}
These metrics are widely adopted in the prognostics community as baseline measures of model performance~\cite{r39,r42}.

\subsubsection{NASA Prognostics Score}

The NASA Prognostics Scoring Function~\cite{r27} applies exponentially asymmetric penalties to prediction errors:
\begin{equation}
  S = \sum_{i=1}^{N} s_i, \qquad
  s_i =
  \begin{cases}
    e^{-d_i/13} - 1, & \text{if } d_i < 0,\\[2pt]
    e^{d_i/10} - 1,  & \text{if } d_i \geq 0,
  \end{cases}
\end{equation}
where $d_i = \hat{y}_i - y_i$. Late predictions ($d_i < 0$, underestimating true RUL) receive a steeper penalty than early predictions ($d_i > 0$), directly capturing the safety-critical nature of prognostics. This scoring function has become a \emph{de facto} standard in the C-MAPSS benchmarking literature.

\subsubsection{Alpha-Accuracy}

The $\alpha$-accuracy metric~\cite{r28} reports the percentage of predictions falling within $\alpha\%$ of the true RUL value. We evaluate at $\alpha = 20\%$:
\begin{equation}
  \alpha\text{-}20\% = \frac{1}{N}\sum_{i=1}^{N} \mathbb{1}\!\left(\left|\hat{y}_i - y_i\right| \leq 0.20 \cdot y_i\right) \times 100\%.
\end{equation}
This metric is particularly informative for maintenance planning, as it indicates the proportion of predictions accurate enough for operational decision-making thresholds.

\subsubsection{Directional Imbalance}

The Directional Imbalance (DI) quantifies the degree of asymmetry between early and late predictions:
\begin{equation}
  \mathrm{DI} = \left|\mathrm{ER} - \mathrm{LR}\right|,
\end{equation}
where the Early Rate and Late Rate are defined as
\begin{equation}
  \mathrm{ER} = \frac{N_{\text{early}}}{N} \times 100\%, \qquad
  \mathrm{LR} = \frac{N_{\text{late}}}{N} \times 100\%,
\end{equation}
with $N_{\text{early}}$ and $N_{\text{late}}$ denoting the number of predictions that overestimate and underestimate the true RUL, respectively. A DI of 0\% indicates a perfectly balanced early/late prediction profile, while higher values reflect greater directional asymmetry. This metric is essential for assessing whether multi-objective optimization achieves better calibration of prediction timeliness compared to single-objective alternatives, as directional preferences have direct implications for maintenance scheduling decisions~\cite{r28}.

\subsection{Experimental Setup}

\begin{figure}[t]
  \centering
  \includegraphics[width=0.78\textwidth]{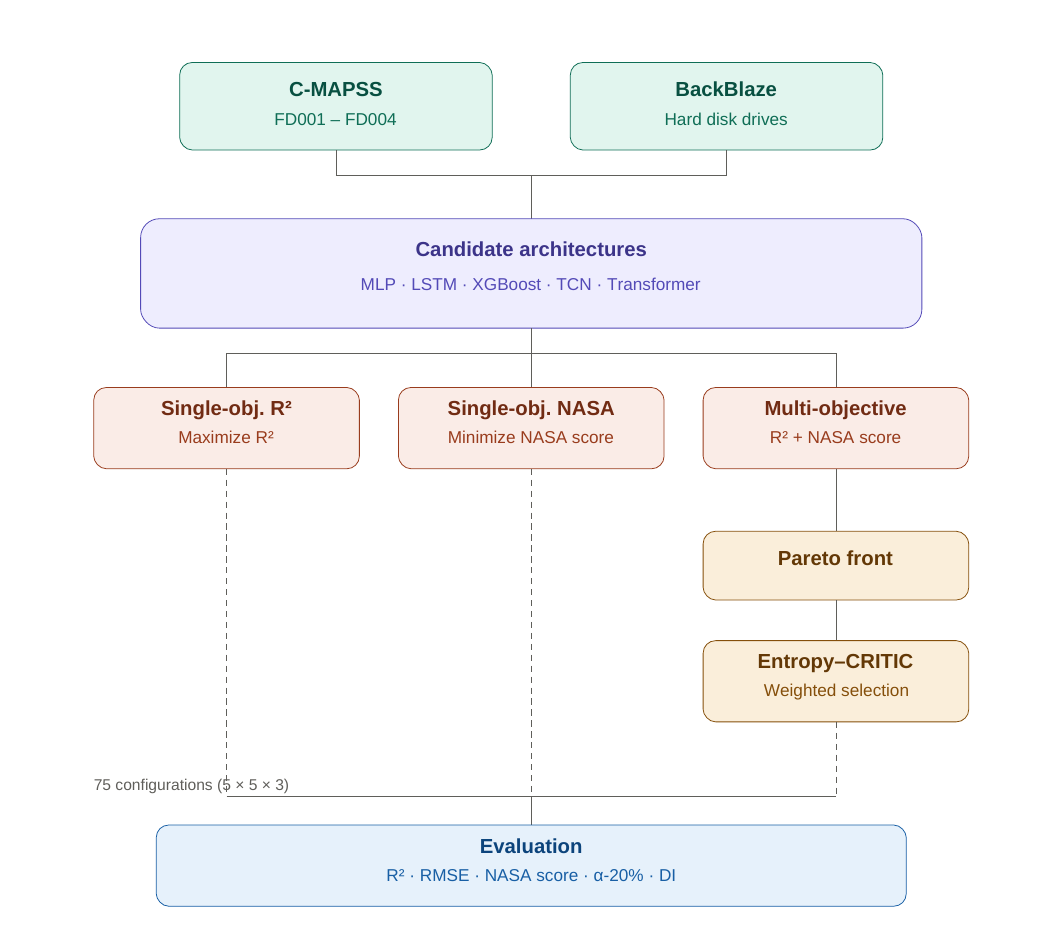}
  \caption{The methodology of the study.}
  \label{fig:method}
\end{figure}

Each algorithm is evaluated across all five datasets under each of the three optimization strategies, yielding a total of 75 experimental configurations (5 algorithms $\times$ 5 datasets $\times$ 3 strategies). The hyperparameter search is conducted using 30 Optuna trials per configuration with the TPE sampler for single-objective strategies and NSGA-II for multi-objective optimization. An early stopping mechanism terminates the search if no improvement is observed for 15 consecutive trials, subject to a minimum of 10 completed trials. The complete hyperparameter search spaces for each algorithm are provided in \Cref{app:hpo}; dimensionality ranges from 5 parameters (LSTM) to 11 parameters (MLP, XGBoost, TCN, Transformer). For C-MAPSS, the datasets are split into training, validation, and test partitions following standard protocols~\cite{r24}, with the piecewise linear RUL target capped at 125 cycles~\cite{r38,r39} and sensor selection retaining 14 informative channels after excluding constant-variance and highly correlated sensors. For BackBlaze, daily SMART attribute readings are processed following Xu et al.~\cite{r26}, with the same temporal train--validation--test split structure to maintain consistency across benchmarks. All preprocessing transformations are fitted exclusively on training data to prevent information leakage. Reported metrics are computed on held-out test sets, with the validation set used exclusively for hyperparameter selection and early stopping. For multi-objective runs, the final configuration is selected from the NSGA-II Pareto front using Entropy--CRITIC weighting~\cite{r22}. All experiments are seeded (seed = 42) for reproducibility. \Cref{fig:method} displays the methodology of the current study.

\section{Experimental Results}

\subsection{Effect of the Optimization Strategy}

\Cref{tab:strategy} reports the performance of each optimization strategy across all five datasets, averaged over all five algorithms. This aggregation isolates the effect of the optimization objective from architecture-specific variation.

\begin{table}[t]
  \centering
  \caption{Comparative analysis across the HPO strategies (averaged over the five algorithms).}
  \label{tab:strategy}
  \small
  \begin{tabular}{@{}llrrrrr@{}}
    \toprule
    \textbf{Dataset} & \textbf{Strategy} & \textbf{\Rsq{}} & \textbf{RMSE} & \textbf{NASA} & \textbf{$\bm{\alpha}$-20\%} & \textbf{DI} \\
    \midrule
    \multirow{3}{*}{C-MAPSS FD001} & Single-\Rsq{}   & 0.890 & 13.19 & 255.8   & 70.0\% & 6.20 \\
                                   & Single-NASA     & 0.889 & 13.32 & 248.2   & 73.6\% & 4.20 \\
                                   & Multi-objective & 0.878 & 13.87 & 305.8   & 66.2\% & 4.80 \\
    \midrule
    \multirow{3}{*}{C-MAPSS FD002} & Single-\Rsq{}   & 0.903 & 13.34 & 795.6   & 65.6\% & 9.15 \\
                                   & Single-NASA     & 0.889 & 14.28 & 960.4   & 60.8\% & 13.01 \\
                                   & Multi-objective & 0.889 & 14.21 & 928.1   & 65.3\% & 6.76 \\
    \midrule
    \multirow{3}{*}{C-MAPSS FD003} & Single-\Rsq{}   & 0.879 & 13.49 & 378.8   & 67.8\% & 13.40 \\
                                   & Single-NASA     & 0.894 & 12.75 & 328.8   & 72.8\% & 11.80 \\
                                   & Multi-objective & 0.894 & 12.71 & 306.4   & 71.2\% & 8.60 \\
    \midrule
    \multirow{3}{*}{C-MAPSS FD004} & Single-\Rsq{}   & 0.881 & 14.83 & 1,281.5 & 64.9\% & 7.50 \\
                                   & Single-NASA     & 0.880 & 14.86 & 1,162.7 & 64.7\% & 7.42 \\
                                   & Multi-objective & 0.883 & 14.66 & 1,221.9 & 66.6\% & 4.11 \\
    \midrule
    \multirow{3}{*}{BackBlaze}     & Single-\Rsq{}   & 0.335 & 13.73 & 55,431  & 24.7\% & 1.45 \\
                                   & Single-NASA     & 0.221 & 11.77 & 27,641  & 21.7\% & 3.39 \\
                                   & Multi-objective & 0.258 & 12.43 & 32,133  & 22.5\% & 2.09 \\
    \bottomrule
  \end{tabular}
\end{table}

Across all four C-MAPSS sub-datasets, the three strategies produce near-identical \Rsq{} values, with aggregated averages of 0.889 (Single-\Rsq{}), 0.888 (Single-NASA), and 0.886 (Multi-Objective). This convergence confirms that the choice of optimization objective has a negligible effect on explained variance within the controlled simulation environment---a finding that challenges the assumption that maximizing a single accuracy metric is sufficient for prognostic model development~\cite{r18}.

The critical distinction emerges in directional imbalance. Multi-objective optimization consistently achieves the lowest DI across all C-MAPSS sub-datasets: 4.80 on FD001, 6.76 on FD002, 8.60 on FD003, and 4.11 on FD004. Averaged across sub-datasets, this represents a 33\% reduction compared to both Single-\Rsq{} (9.06) and Single-NASA (9.11), consistent with the theoretical expectation that Pareto-based search explores a broader hyperparameter region~\cite{r13,r14}. Crucially, this advantage scales with dataset complexity: from a 23\% reduction on FD001 (4.80 vs.\ 4.20) to 45\% on FD004 (4.11 vs.\ 7.42), suggesting that single-objective optimizers increasingly converge to locally optimal configurations as the objective landscape becomes more rugged. NSGA-II's population-based exploration mitigates this by maintaining diverse solution candidates~\cite{r15}.

On BackBlaze, the strategies diverge substantially. Single-\Rsq{} achieves the highest \Rsq{} (0.335), followed by Multi-Objective (0.258) and Single-NASA (0.221)---a spread of 0.114 compared to just 0.003 on C-MAPSS. This reveals that the two objectives become strongly conflicting on noisy data: Single-NASA reduces the NASA Score by 50\% relative to Single-\Rsq{} (27,641 vs.\ 55,431) but at a 34\% cost in \Rsq{}. Multi-objective optimization positions itself between the two extremes, achieving a NASA Score (32,133) comparable to Single-NASA while retaining higher \Rsq{}. The NASA Score itself exhibits extreme instability, with aggregate values two orders of magnitude larger than the C-MAPSS range (248--1,282), consistent with the observation by Ferreira and Gon\c{c}alves~\cite{r1} that performance metrics become increasingly unstable on real-world degradation data. The DI pattern reverses on BackBlaze---Single-\Rsq{} produces the lowest DI (1.45)---but this reflects the near-random nature of predictions rather than improved calibration: when a model explains little variance, errors distribute symmetrically, producing naturally balanced but uninformative splits. This confirms that the multi-objective DI advantage is meaningful only when the model has sufficient predictive power to exhibit systematic directional tendencies.

The Entropy--CRITIC weights provide direct insight into the objective relationship. On C-MAPSS, weights converge to nearly equal for most algorithm--dataset combinations, confirming that the objectives are co-aligned and explaining the \Rsq{} convergence across strategies~\cite{r13}. On BackBlaze, the weight structure shifts dramatically in architecture-dependent directions: TCN and Transformer assign 0.68--0.70 weight to the NASA Score, prioritizing penalty control; LSTM shows a moderate NASA-leaning pattern (0.38/0.62); XGBoost remains the most balanced (0.46/0.54); and MLP assigns 0.75 weight to \Rsq{}---the only architecture where accuracy dominates selection. These divergent patterns indicate that the relationship between objectives shifts from complementary to conflicting in architecture-specific ways, supporting the argument that multi-objective HPO is most valuable when the relationships between objectives are unknown a priori~\cite{r13,r14}.

\subsection{Algorithm Comparison}

\Cref{tab:algorithms} presents the performance of each algorithm on each dataset, averaged across the three optimization strategies, isolating architectural effects from strategy-specific variation.

\begin{table}[t]
  \centering
  \caption{Comparative analysis across algorithms (averaged over the three optimization strategies).}
  \label{tab:algorithms}
  \small
  \begin{tabular}{@{}llrrrrr@{}}
    \toprule
    \textbf{Dataset} & \textbf{Algorithm} & \textbf{\Rsq{}} & \textbf{RMSE} & \textbf{NASA} & \textbf{$\bm{\alpha}$-20\%} & \textbf{DI} \\
    \midrule
    \multirow{5}{*}{FD001} & XGBoost     & 0.916 & 11.57 & 190.3   & 75.0\% & 1.33 \\
                           & MLP         & 0.899 & 12.73 & 249.1   & 72.0\% & 5.00 \\
                           & LSTM        & 0.891 & 13.23 & 250.8   & 72.0\% & 5.33 \\
                           & Transformer & 0.868 & 14.54 & 348.7   & 69.7\% & 7.67 \\
                           & TCN         & 0.855 & 15.24 & 310.9   & 61.0\% & 6.00 \\
    \midrule
    \multirow{5}{*}{FD002} & XGBoost     & 0.911 & 12.80 & 649.4   & 68.6\% & 4.70 \\
                           & MLP         & 0.907 & 13.09 & 746.9   & 66.5\% & 11.52 \\
                           & LSTM        & 0.906 & 13.13 & 792.6   & 66.4\% & 3.28 \\
                           & Transformer & 0.880 & 14.88 & 1,199.0 & 58.9\% & 18.21 \\
                           & TCN         & 0.864 & 15.80 & 1,085.7 & 58.8\% & 10.49 \\
    \midrule
    \multirow{5}{*}{FD003} & XGBoost     & 0.907 & 11.92 & 211.9   & 70.7\% & 15.33 \\
                           & MLP         & 0.905 & 12.06 & 225.3   & 76.3\% & 10.67 \\
                           & LSTM        & 0.892 & 12.84 & 392.3   & 74.3\% & 2.00 \\
                           & Transformer & 0.884 & 13.31 & 349.1   & 67.7\% & 16.67 \\
                           & TCN         & 0.856 & 14.79 & 511.2   & 64.0\% & 11.67 \\
    \midrule
    \multirow{5}{*}{FD004} & LSTM        & 0.891 & 14.19 & 1,032.2 & 62.5\% & 1.88 \\
                           & XGBoost     & 0.885 & 14.58 & 1,005.8 & 68.7\% & 3.09 \\
                           & MLP         & 0.885 & 14.60 & 1,116.5 & 67.5\% & 7.26 \\
                           & Transformer & 0.881 & 14.84 & 1,562.7 & 66.8\% & 9.41 \\
                           & TCN         & 0.866 & 15.72 & 1,392.9 & 61.6\% & 10.08 \\
    \midrule
    \multirow{5}{*}{BackBlaze} & XGBoost     & 0.301 & 12.39 & 33,986 & 23.4\% & 1.66 \\
                               & MLP         & 0.296 & 12.47 & 45,477 & 23.6\% & 2.44 \\
                               & TCN         & 0.276 & 12.60 & 34,510 & 22.1\% & 0.61 \\
                               & LSTM        & 0.252 & 12.80 & 36,918 & 23.0\% & 3.99 \\
                               & Transformer & 0.235 & 12.97 & 41,118 & 22.9\% & 2.85 \\
    \bottomrule
  \end{tabular}
\end{table}

On FD001, FD002, and FD003---sub-datasets with simpler operating conditions---XGBoost ranks first on \Rsq{} (0.916, 0.911, 0.907), RMSE, and NASA Score, with the best $\alpha$-20\% and DI on FD001. These findings align with Alfarizi et al.~\cite{r8} and reinforce the observation that ensemble methods often outperform deep architectures on tabular sensor data when the signal-to-noise ratio is favorable~\cite{r31}.

The hierarchy shifts on FD004, where LSTM achieves the highest \Rsq{} (0.891), lowest RMSE (14.19), and lowest DI (1.88), overtaking XGBoost (\Rsq{} = 0.885, DI = 3.09). This suggests that the gating mechanism can separate condition-dependent variation from fault progression---a capacity a memoryless tree ensemble lacks---consistent with Gugulothu et al.~\cite{r6} and Thoppil et al.~\cite{r12}. Notably, LSTM maintains the lowest DI across all C-MAPSS sub-datasets (1.33--5.33), indicating that its recurrent dynamics intrinsically regularize prediction direction.

TCN and Transformer consistently occupy the bottom two positions across all C-MAPSS sub-datasets, contrasting with recent reports of their effectiveness in aero-engine prognostics~\cite{r2,r4,r9}. TCN yields the lowest \Rsq{} on all four subsets, while the Transformer produces the highest NASA Scores on FD002 (1,199.0) and FD004 (1,562.7) and the highest DI of any combination in the study (18.21 on FD002). These results suggest that the additional capacity of attention-based and dilated convolutional architectures does not automatically translate into improved prognostic performance, likely due to the relatively short sequences and limited training samples in C-MAPSS~\cite{r3,r7}. The discrepancy with cited studies likely reflects differences in feature engineering, training protocols, and optimization strategy---reinforcing the central thesis that optimization configuration shapes prognostic outcomes as much as architecture.

The transition to BackBlaze reveals a pronounced performance collapse: \Rsq{} drops from 0.86--0.92 to 0.24--0.30 and $\alpha$-20\% from 61--75\% to 22--24\%. The algorithm ranking undergoes a configurational reversal---TCN rises from last to third (\Rsq{} = 0.276), LSTM drops to fourth (0.252), and the Transformer remains last on both benchmarks. The performance spread compresses from 0.061 (FD001) to 0.066 (BackBlaze) with substantially less practical differentiation, indicating that dataset noise increasingly dominates model-specific differences~\cite{r33}.

Finally, multi-objective optimization does not affect all architectures uniformly. On C-MAPSS, TCN and MLP benefit most from Pareto-based search (DI reductions of 76\% and 43\%), while LSTM maintains low DI regardless of strategy (2.20--3.59) and the Transformer exhibits persistently high DI (10.97--15.17). For XGBoost, MLP, and LSTM, multi-objective optimization simultaneously achieves the highest C-MAPSS \Rsq{} (0.907, 0.901, 0.901), indicating no accuracy cost for well-behaved architectures. For TCN and Transformer, however, it produces the lowest \Rsq{} (0.855 and 0.865). This asymmetric interaction implies that the optimal pairing of architecture and optimization strategy depends on the model's intrinsic directional tendencies~\cite{r13}.

\section{Conclusion}

This study compared five architectures (MLP, LSTM, XGBoost, TCN, Transformer) for RUL prediction under single-objective \Rsq{}, single-objective NASA Score, and multi-objective NSGA-II optimization with Entropy--CRITIC selection across 75 configurations on C-MAPSS and BackBlaze. Three findings emerged. First, the optimization objective reshapes prediction behavior more than it affects aggregate accuracy: all strategies converged to near-identical \Rsq{} on C-MAPSS, yet multi-objective optimization reduced directional imbalance by 33\%, with this advantage scaling with dataset complexity. Second, the transition to BackBlaze revealed a pronounced generalization gap, NASA Score instability, and strategy divergence---confirmed by Entropy--CRITIC weights that were balanced on C-MAPSS but diverged in architecture-dependent directions on BackBlaze, indicating that the relationship between objectives shifts from complementary to conflicting as data complexity increases. Third, architecture rankings proved configuration-dependent: XGBoost dominated simpler conditions, LSTM overtook it under complex multi-condition scenarios with the most balanced directional profile, while on BackBlaze the rankings reversed entirely.

Multi-objective HPO with Entropy--CRITIC selection provides a principled procedure for obtaining directionally balanced RUL predictions without manual preference articulation~\cite{r13,r14}. The 33\% DI reduction translates into fewer unexpected breakdowns from systematically late predictions without a compensating increase in premature replacements, narrowing the gap between planned and actual maintenance timing. The Entropy--CRITIC weight structure further serves as a deployment diagnostic: balanced weights indicate that single-metric tuning is adequate, whereas divergent weights signal that multi-objective search is necessary---a distinction the performance collapse on BackBlaze underscores, cautioning against uncritical adoption of simulation-derived objectives without domain-specific validation~\cite{r33}. The configurational dependence of algorithm ranking implies that architecture selection should be guided by operational complexity rather than single-benchmark performance~\cite{r1,r31}, while the dominance of simpler models reinforces that well-tuned lower-complexity architectures remain competitive on structured degradation data of moderate scale~\cite{r8}.

Several limitations apply. The fixed computational budget may have disadvantaged deeper architectures with larger search spaces~\cite{r18}. The multi-objective framework was restricted to two objectives, excluding model complexity, inference latency, and calibration quality. BackBlaze represents a single industrial domain. Each architecture was evaluated independently without hybrid or ensemble combinations. The Entropy--CRITIC mechanism is a static post-hoc procedure that does not incorporate domain expert preferences.

Future work should expand optimization budgets to determine whether deeper architectures can close the performance gap~\cite{r10}, incorporate additional objectives for resource-constrained deployment~\cite{r13}, and extend evaluation to manufacturing, energy, and transportation domains~\cite{r29,r36}. Hybrid approaches, adaptive Pareto selection integrating domain expertise~\cite{r14,r15}, and probabilistic RUL frameworks including survival analysis and conformal prediction offer promising directions beyond the single-point paradigm examined here~\cite{r3}.

\section*{Acknowledgments}

The authors acknowledge the use of Claude 4.6 Opus (Anthropic) for writing assistance and code development support during the preparation of this manuscript.

\clearpage
\appendix
\section{Hyperparameter Search Spaces}
\label{app:hpo}

\begin{table}[!h]
  \centering
  \caption{Hyperparameter search spaces for each algorithm.}
  \label{tab:hpo}
  \small
  \begin{tabular}{@{}llll@{}}
    \toprule
    \textbf{Algorithm} & \textbf{Hyperparameter} & \textbf{Search range} & \textbf{Type} \\
    \midrule
    \multirow{11}{*}{MLP}
      & Sequence length      & \{20, 30, 40, 50\}            & Categorical \\
      & Hidden layers        & [2, 5]                        & Integer \\
      & First layer size     & \{128, 256, 512\}             & Categorical \\
      & Layer shrink factor  & \{1.0, 0.75, 0.5\}            & Categorical \\
      & Dropout rate         & [0.1, 0.5], step 0.1          & Float \\
      & L2 regularization    & [$10^{-5}$, $10^{-3}$]        & Log-float \\
      & Batch normalization  & \{True, False\}               & Categorical \\
      & Activation           & \{ReLU, ELU, SELU\}           & Categorical \\
      & Model variant        & \{flatten, stats\}            & Categorical \\
      & Batch size           & \{64, 128, 256\}              & Categorical \\
      & Learning rate        & [$10^{-4}$, $10^{-2}$]        & Log-float \\
    \midrule
    \multirow{5}{*}{LSTM}
      & Sequence length      & \{30, 50\}                    & Categorical \\
      & First LSTM units     & \{64, 128\}                   & Categorical \\
      & Bidirectional        & \{True, False\}               & Categorical \\
      & Batch size           & \{128, 256\}                  & Categorical \\
      & Learning rate        & [$5\times10^{-4}$, $5\times10^{-3}$] & Log-float \\
    \midrule
    \multirow{11}{*}{XGBoost}
      & Sequence length      & \{20, 30, 40, 50\}            & Categorical \\
      & Model variant        & \{flatten, stats\}            & Categorical \\
      & Number of estimators & [100, 2000], step 100         & Integer \\
      & Max depth            & [3, 12]                       & Integer \\
      & Learning rate        & [0.005, 0.3]                  & Log-float \\
      & Subsample ratio      & [0.5, 1.0], step 0.05         & Float \\
      & Col.\ sample by tree & [0.3, 1.0], step 0.05         & Float \\
      & Min child weight     & [1, 20]                       & Integer \\
      & Gamma                & [0.0, 5.0], step 0.1          & Float \\
      & L1 reg.\ ($\alpha$)  & [$10^{-5}$, 10.0]             & Log-float \\
      & L2 reg.\ ($\lambda$) & [$10^{-5}$, 10.0]             & Log-float \\
    \midrule
    \multirow{11}{*}{TCN}
      & Sequence length      & \{20, 30, 40, 50\}            & Categorical \\
      & Model variant        & \{sequence, stats\}           & Categorical \\
      & Number of filters    & \{32, 64, 128\}               & Categorical \\
      & Kernel size          & \{2, 3, 4, 5\}                & Categorical \\
      & TCN blocks           & [2, 6]                        & Integer \\
      & Dropout rate         & [0.05, 0.4], step 0.05        & Float \\
      & L2 regularization    & [$10^{-6}$, $10^{-3}$]        & Log-float \\
      & Batch normalization  & \{True, False\}               & Categorical \\
      & Activation           & \{ReLU, ELU, SELU\}           & Categorical \\
      & Batch size           & \{64, 128, 256\}              & Categorical \\
      & Learning rate        & [$10^{-4}$, $10^{-2}$]        & Log-float \\
    \midrule
    \multirow{11}{*}{Transformer}
      & Sequence length      & \{20, 30, 40, 50\}            & Categorical \\
      & Model variant        & \{sequence, stats\}           & Categorical \\
      & Attention heads      & \{2, 4, 8\}                   & Categorical \\
      & Model dimension      & \{32, 64, 128, 256\}          & Categorical \\
      & Encoder layers       & [1, 4]                        & Integer \\
      & FFN dim.\ multiplier & \{2, 4\}                      & Categorical \\
      & Dropout rate         & [0.05, 0.4], step 0.05        & Float \\
      & Positional encoding  & \{True, False\}               & Categorical \\
      & Activation           & \{ReLU, GELU\}                & Categorical \\
      & Batch size           & \{64, 128, 256\}              & Categorical \\
      & Learning rate        & [$10^{-4}$, $5\times10^{-3}$] & Log-float \\
    \bottomrule
  \end{tabular}
\end{table}

\end{document}